\documentclass[sigconf]{acmart}

\usepackage{amsmath,amsfonts}
\usepackage{algorithmic}
\usepackage{algorithm}
\usepackage{array}
\usepackage{subfig}
\usepackage{textcomp}
\usepackage{stfloats}
\usepackage{url}
\usepackage{verbatim}
\usepackage{graphicx}
\usepackage{multirow}

\AtBeginDocument{%
  \providecommand\BibTeX{{%
    \normalfont B\kern-0.5em{\scshape i\kern-0.25em b}\kern-0.8em\TeX}}}

\copyrightyear{2022}
\acmYear{2022}
\setcopyright{acmcopyright}\acmConference[SIGIR '22]{Proceedings of the 45th International ACM SIGIR Conference on Research and Development in Information Retrieval}{July 11--15, 2022}{Madrid, Spain}
\acmBooktitle{Proceedings of the 45th International ACM SIGIR Conference on Research and Development in Information Retrieval (SIGIR '22), July 11--15, 2022, Madrid, Spain}
\acmPrice{15.00}
\acmDOI{10.1145/3477495.3532064}
\acmISBN{978-1-4503-8732-3/22/07}

\begin{document}
\fancyhead{}
\title{Tag-assisted Multimodal Sentiment Analysis under Uncertain Missing Modalities}


\author{Jiandian Zeng}
\affiliation{%
  \institution{State Key Laboratory of IoT for Smart City, University of Macau}
  \city{Macau}
  \country{China}}
\email{yb87470@um.edu.mo}

\author{Tianyi Liu}
\affiliation{%
  \institution{Shanghai Jiao Tong University}
  \city{Shanghai}
  \country{China}}
\email{liutianyi@sjtu.edu.cn}

\author{Jiantao Zhou}
\authornote{Corresponding Author}
\affiliation{%
  \institution{State Key Laboratory of IoT for Smart City, University of Macau}
  \city{Macau}
  \country{China}}
\email{jtzhou@um.edu.mo}


\begin{abstract}
  Multimodal sentiment analysis has been studied under the assumption that all modalities are available. However, such a strong assumption does not always hold in practice, and most of multimodal fusion models may fail when partial modalities are missing. Several works have addressed the missing modality problem; but most of them only considered the single modality missing case, and ignored the practically more general cases of multiple modalities missing. To this end, in this paper, we propose a Tag-Assisted Transformer Encoder (TATE) network to handle the problem of missing uncertain modalities. Specifically, we design a tag encoding module to cover both the single modality and multiple modalities missing cases, so as to guide the network's attention to those missing modalities. Besides, we adopt a new space projection pattern to align common vectors. Then, a Transformer encoder-decoder network is utilized to learn the missing modality features. At last, the outputs of the Transformer encoder are used for the final sentiment classification. Extensive experiments are conducted on CMU-MOSI and IEMOCAP datasets, showing that our method can achieve significant improvements compared with several baselines.
\end{abstract}





\maketitle

\section{Introduction}

\begin{figure}[t]
    \centering
    \includegraphics[width=0.85\linewidth ]{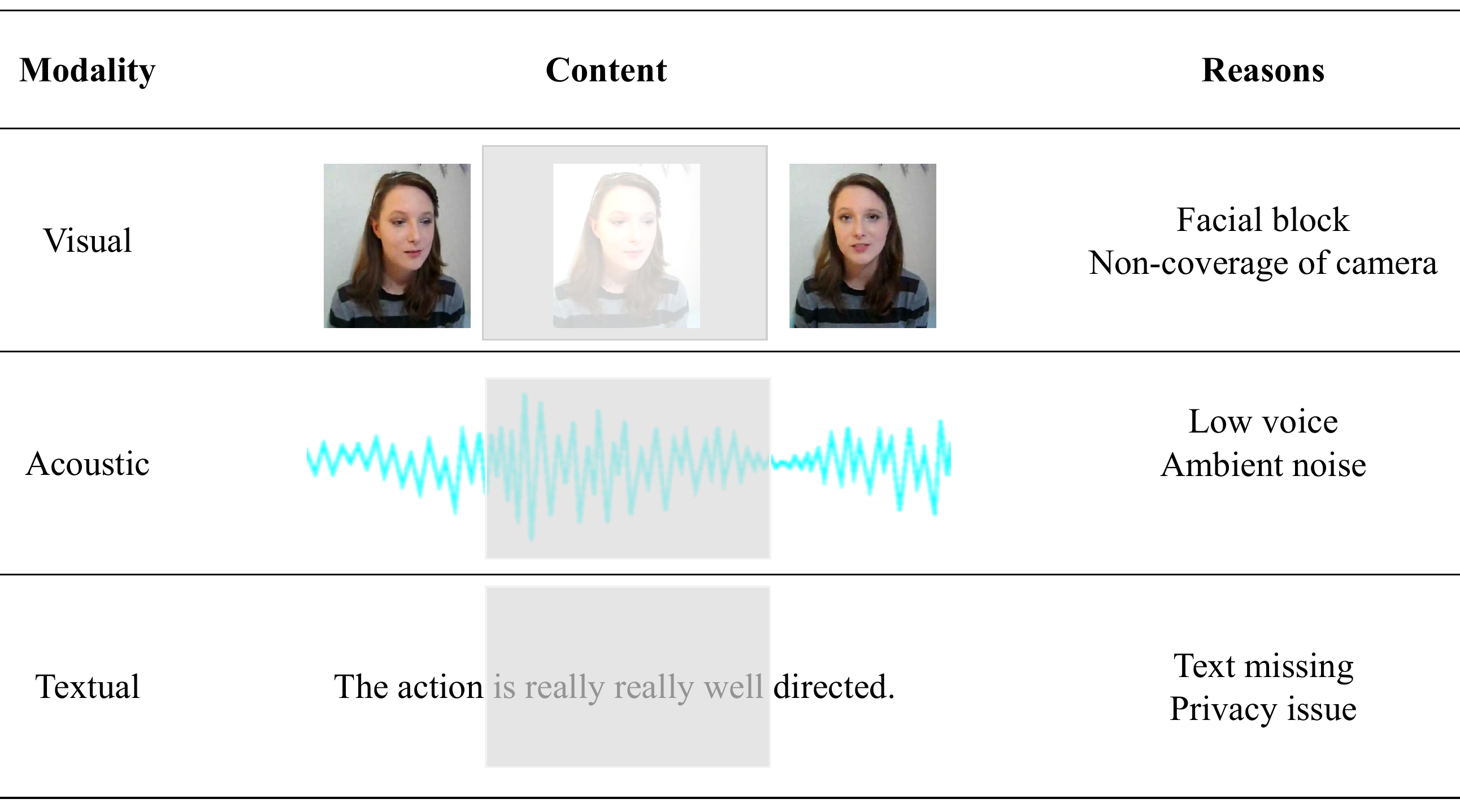}
    \caption{Examples of missing modalities.}
    \label{fig1}
\end{figure}

Nowadays, sentiment analysis has attracted intensive interest in extracting human's emotion and opinion~\cite{wang2019investigating, zhang2020knowledge}, among which multimodal sentiment analysis is becoming an especially popular research direction with the massive amounts of online content. Besides, it has been shown that combining different modalities can learn complementary features, resulting in better joint multimodal representations~\cite{shraga2020web, springstein2021quti}. Most prior works on multimodal fusion~\cite{chen2017multimodal, xu2018co, xu2020social} assumed that all modalities are always available when training and testing. However, in real life, we often encounter scenarios that partial modalities could be missing. For example, as shown in Fig.~\ref{fig1}, the visual features may be blocked due to the non-coverage of camera; the acoustic information may be unavailable due to the enormous ambient noise; and the textual information may be absent due to the privacy issue. Therefore, how to handle missing modalities is emerging as a hot topic in the multimodal area.

Previous works~\cite{mazumder2010spectral, shao2013clustering, parthasarathy2020training} simply discarded missing modalities or utilized matrix completion methods to impute missing modalities, and somewhat degraded overall performance. Zhao \textit{et al.}~\cite{shao2013clustering} completed the kernel matrices of the modality using the common instances in different modalities. In~\cite{parthasarathy2020training}, the visual modality was ablated when training with missing data. Owing to the strong learning ability of deep learning, recent works have employed neural networks to learn latent relationships among available modalities. To tackle the missing modality problem, Tran \textit{et al.}~\cite{tran2017missing} first identified the general problem of missing modality in multimodal data, and proposed a Cascaded Residual Auto-encoder (CRA) network to learn complex relationship from different modalities. More recently, Zhao \textit{et al.}~\cite{zhao2021missing} adopted cycle consistency learning with CRA to recover missing modalities. Yuan \textit{et al.}~\cite{yuan2021transformer} designed a Transformer-based feature reconstruction network to guide the extractor in obtaining the semantics of missing modality features. However, most of the above works all assumed that there is only one missing modality, and ignored the practically more general cases of multiple modalities missing. That is, they require training a new model to fit each missing modality case, which is both costly and inconvenient. In reality, the pattern of missing modalities could be uncertain, e.g., one or two modalities are randomly absent. To tackle the above issues, two challenges should be addressed: 1) will the model still work when multiple modalities are absent? and 2) how to learn robust joint representations when the missing modalities are uncertain?

In this paper, we propose a \textbf{T}ag-\textbf{A}ssisted \textbf{T}ransformer \textbf{E}ncoder (TATE) network to learn complementary features among modalities. For the first challenge, we design a tag encoding module to mark missing modalities, aiming to direct the network's attention to absent modalities. As will be shown later, the attached tag not only can cover both the single modality and multiple modalities absent situations, but also can assist in joint representation learning. For the second challenge, we first adopt the Transformer~\cite{vaswani2017attention} as the extractor to capture intra-modal features, and then apply a two-by-two projection pattern to map them into a common space. After that, the pre-trained network trained with full modalities is utilized to supervise the encoded vectors. At last, the outputs generated by a Transformer encoder are fed into a classifier for sentiment prediction. Our contributions are summarized as follows:

\begin{itemize}
    \item We propose the TATE network to handle the multiple modalities missing problem for multimodal sentiment analysis. The code is publicly available$\footnote{https://github.com/JaydenZeng/TATE}$.
    \item We design a tag encoding module to cover both the single modality and multiple modalities absent situations, and adopt a new common space projection module to learn joint representations.
    \item Our proposed model TATE achieves significant improvements compared with several benchmarks on CMU-MOSI and IEMOCAP datasets, validating the superiority of our model.
\end{itemize}

\section{Related Works}~\label{related_work}
In this section, we first introduce the concept of multimodal sentiment analysis, and then review the related methods of handing missing modalities.

\subsection{Multimodal Sentiment Analysis}
As a core branch of sentiment analysis~\cite{zhuang2020joint, geng2021iterative}, multimodal sentiment analysis has attracted significant attention in recent years~\cite{poria2018multimodal, yu2019entity,  mai2021analyzing, stappen2021sentiment}. Compared to a single modality case, multimodal sentiment analysis is more challenging due to the complexity of handling and analyzing data from different modalities. 

To learn joint representations of multimodal, three multimodal fusion strategies are applied: 1) \textit{early fusion} directly combines features of different modalities before the classification. Majumder \textit{et al.}~\cite{majumder2018multimodal} proposed a hierarchical fusion strategy to fuse acoustic, visual and textual modalities, and proved the effectiveness of two-by-two fusion pattern; 2) \textit{late fusion} adopts the average score of each modality as the final weights. Guo \textit{et al.}~\cite{guo2017online} adopted an online early-late fusion scheme to explore complementary relationship for the sign language recognition, where late fusion further aggregated features combined by the early fusion; and 3) \textit{intermediate fusion} utilizes a shared layer to fuse features. Xu \textit{et al.}~\cite{xu2020reasoning} constructed the decomposition and relation networks to represent the commonality and discrepancy among modalities. Hazarika \textit{et al.}~\cite{hazarika2020misa} designed a multimodal learning framework that can learn modality-invariant and modality-specific representations by projecting each modality into two distinct sub-spaces. However, few of the above multimodal fusion models can handle the cases when partial modalities are missing.

\subsection{Missing Modalities Methods}
In recent years, many works focused on handing the missing modality problem, and they can be generally categorized into two groups: 1) generative methods~\cite{baldi2012autoencoders, tran2017missing, du2018semi, shang2017vigan, zhang2020deep}; and 2) joint learning methods~\cite{pham2019found, wang2020transmodality, zhao2021missing, yuan2021transformer}.

Generative methods learn to generate new data with similar distributions to obey the distribution of the observed data. With the ability to learn latent representations, the auto-encoder (AE)~\cite{baldi2012autoencoders} is widely used. Vincent \textit{et al.}~\cite{vincent2008extracting} extracted features with AE based on the idea of making the learned representations robust to partial corruption of the input data. Kingma \textit{et al.}~\cite{kingma2014auto} designed a Variational Auto-Encoder (VAE) to infer and learn features with simple ancestral sampling. Besides, inspired by the residual connection network~\cite{he2016deep}, Tran \textit{et al.}~\cite{tran2017missing} proposed a Cascaded Residual Auto-encoder (CRA) to impute data with missing modality, which combined a series of residual AEs into a cascaded architecture to learn relationships among different modalities. As for the Generative Adversarial Networks (GAN)~\cite{goodfellow2014generative}, Shang \textit{et al.}~\cite{shang2017vigan} treated each view as a separate domain, and identified domain-to-domain mappings via a GAN using randomly-sampled data from each view. Besides, the domain mapping technique is also considered to impute missing data.  Cai \textit{et al.}~\cite{cai2018deep} formulated the missing modality problem as a conditional image generation task, and designed a 3D encoder-decoder network to capture modality relations. They also incorporated the available category information during training to enhance the robustness of the model. Moreover, Zhao \textit{et al.}~\cite{zhang2020deep} developed a cross partial multi-view network to model complex correlations among different views, where multiple discriminators are used to generate missing data.

Joint learning methods try to learn joint representations based on the relations among different modalities~\cite{pham2019found, kim2020hypergraph, akbari2021vatt}. Based on the idea that the cycle consistency loss can retain maximal information from all modalities, Pham \textit{et al.}~\cite{pham2019found} investigated learning robust representations via cyclic translations from source to target modalities. Zhao \textit{et al.}~\cite{zhao2021missing} also applied cycle consistency learning for missing modality imputation, where the CRA-based cross-modality imagination module is designed based on paired multimodal data. More recently, Yuan \textit{et al.}~\cite{yuan2021transformer} utilized the Transformer to extract intra-modal and inter-modal relations, and designed a Transformer-based feature reconstruction network to reproduce the semantics of missing modality.

However, most of the above works can only handle the scenarios of missing a single modality, and cannot satisfactorily deal with multiple modalities missing cases since they need to train a new model for each case. As will be clear soon, our works differs the above works in several ways: 1) a tag encoding module is designed to cover all uncertain missing cases; and 2) a new mapping method is applied to learn joint representations in the common space projection module.

\begin{figure*}[t]
    \centering
    \includegraphics[width=0.85\linewidth ]{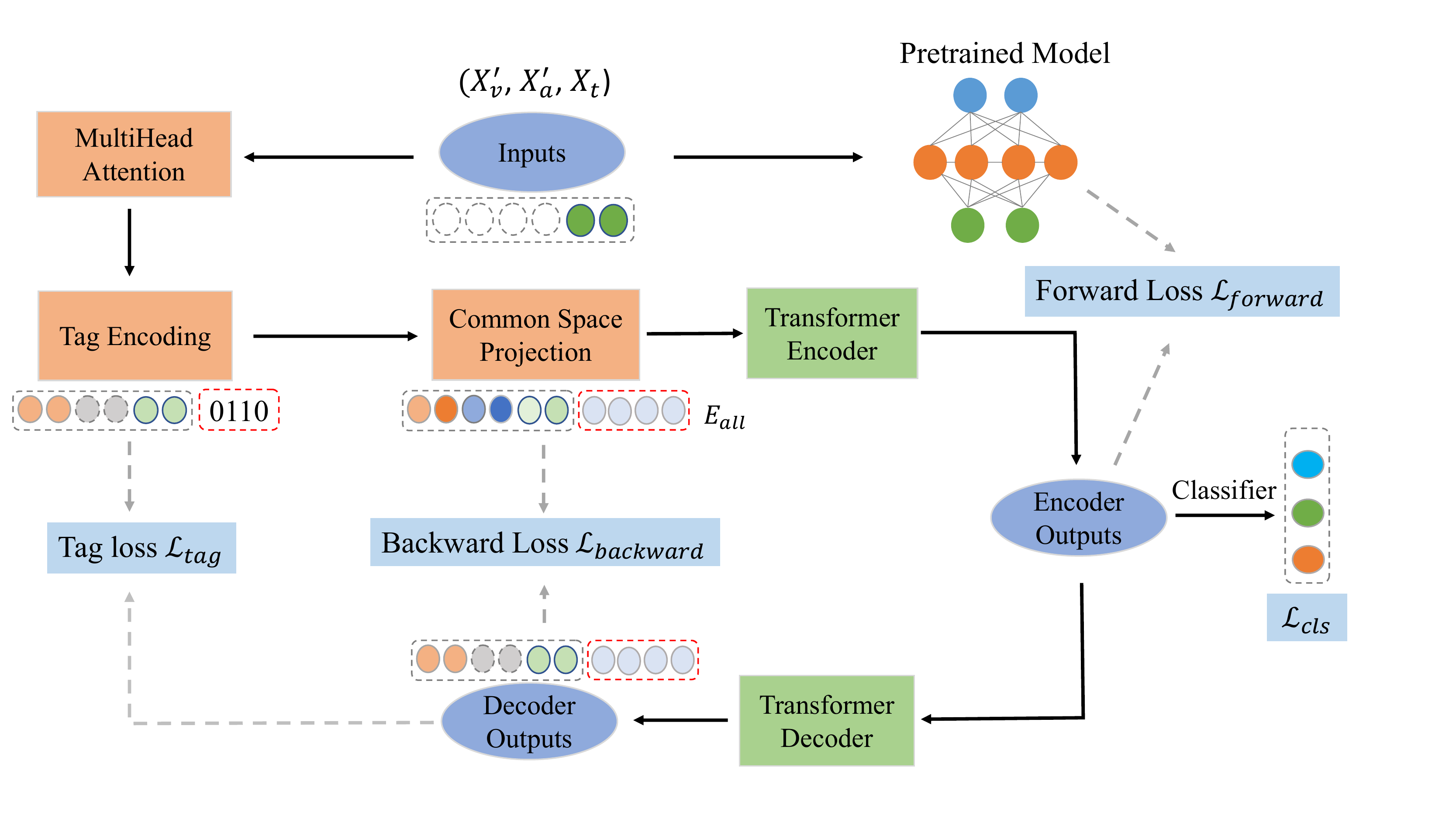}
    \caption{Workflow of the proposed framework. The information flow goes two branches: 1) one goes the pre-trained network, which is trained with full modality data; and 2) another goes to the left multihead attetion module for further encoding.}
    \label{fig2}
\end{figure*}

\section{Methodology}~\label{methodology}
In this section, we first give the problem definition and associated notations. Then, we present the overall workflow of the proposed architecture and the detailed modules.

\subsection{Problem Definition and Notations}
Given a multimodal video segment that contains three modalities: $S =[X_{v}, X_{a}, X_{t}]$, where $X_{v}$, $X_{a}$ and $X_{t}$ denote visual, acoustic and textual modalities respectively. Without loss of generality, we use $X_{m}^{\prime}$ to represent the missing modality, where $m \in \{v, a, t\}$. For instance, assuming that the visual modality and acoustic modality are absent, and the multimodal representation can be denoted as $[X_{v}^{\prime}, X_{a}^{\prime}, X_{t}]$.  The primary task is to classify the overall sentiment (\textit{positive}, \textit{neutral}, or \textit{negative}) under uncertain missing modalities.

\subsection{Overall Framework}
As can be seen in Fig.~\ref{fig2}, the main workflow is as follows: for a given video segment, assuming that the visual modality and acoustic modality are missing, we first mask these missing modalities as 0, and then extract the remaining raw features. Afterwards, the masked multimodal representation goes through two branches: 1) one is encoded by a pre-trained model, which is trained with all full modality data, and 2) another goes through the tag encoding module and the common space projection module to acquire aligned feature vectors. Then, the updated representations are processed by a Transformer encoder, and we calculate the forward similarity loss between the pre-trained vectors and the encoder outputs. Meanwhile, the encoded outputs are fed into a classifier for the sentiment prediction. At last, we compute the backward reconstruction loss and the tag recovery loss to supervise the joint representation learning. Each module will be introduced clearly in following sub-sections.

\subsection{Multi-Head Attention}\label{MultiHead}
Transformer~\cite{vaswani2017attention} not only plays a great role in the Natural Language Processing (NLP) community, but also shows excellent representational capabilities in other areas, such as Computer Vision (CV)~\cite{Chu2021twins}. Instead of using an RNN based structure to capture the sequential information, we employ the Transformer to generate the contextual representation of each modality respectively, where the key component of multi-head dot-product attention can be formalized as follows:

\begin{equation}
    Attention(Q,K,V)=softmax(\frac{QK^{T}}{\sqrt{d}})V,
\end{equation}
where $Q$, $K$ and $V$ are the query, the key, and the value respectively, and $d$ is the dimension of the input.

Instead of utilizing the single attention, the multi-head attention is applied to obtain more information from different semantic spaces:
\begin{equation}
    \begin{aligned}
         E_{M} & = MultiHead(Q,K,V) \\
               & = Concat(head_{1},head_{2}, ..., head_{h})W^{o},
    \end{aligned}
\end{equation}
where $W^{o} \in \mathbb{R}^{d \times d}$ is a weight matrix, $h$ is the head number. Given the input $E$, the $i$-th $head_{i}$ is calculated as follow:
\begin{equation}
head_{i} = Attention(EW_{i}^{Q},EW_{i}^{K},EW_{i}^{V})
\end{equation}
where $W_i^{Q} \in \mathbb{R}^{\frac{d}{h} \times \frac{d}{h}}$, $W_{i}^{K} \in \mathbb{R}^{\frac{d}{h} \times \frac{d}{h}}$ and $W_{i}^{V} \in \mathbb{R}^{\frac{d}{h} \times \frac{d}{h}}$ are the $i$-th weight matrices of the query, the key and the value.

Therefore, the updated modality representations can be formulated as follows:
\begin{equation}
    \begin{aligned}
       &E_{v} = MultiHead(X_{v}^{\prime},X_{v}^{\prime},X_{v}^{\prime}), \\
       & E_{a} = MultiHead(X_{a}^{\prime}, X_{a}^{\prime}, X_{a}^{\prime}), \\
       & E_{t} = MultiHead(X_{t},X_{t},X_{t}).
    \end{aligned}
\end{equation}

\subsection{Tag Encoding}
\begin{figure}[t]
    \centering
    \includegraphics[width=0.95\linewidth ]{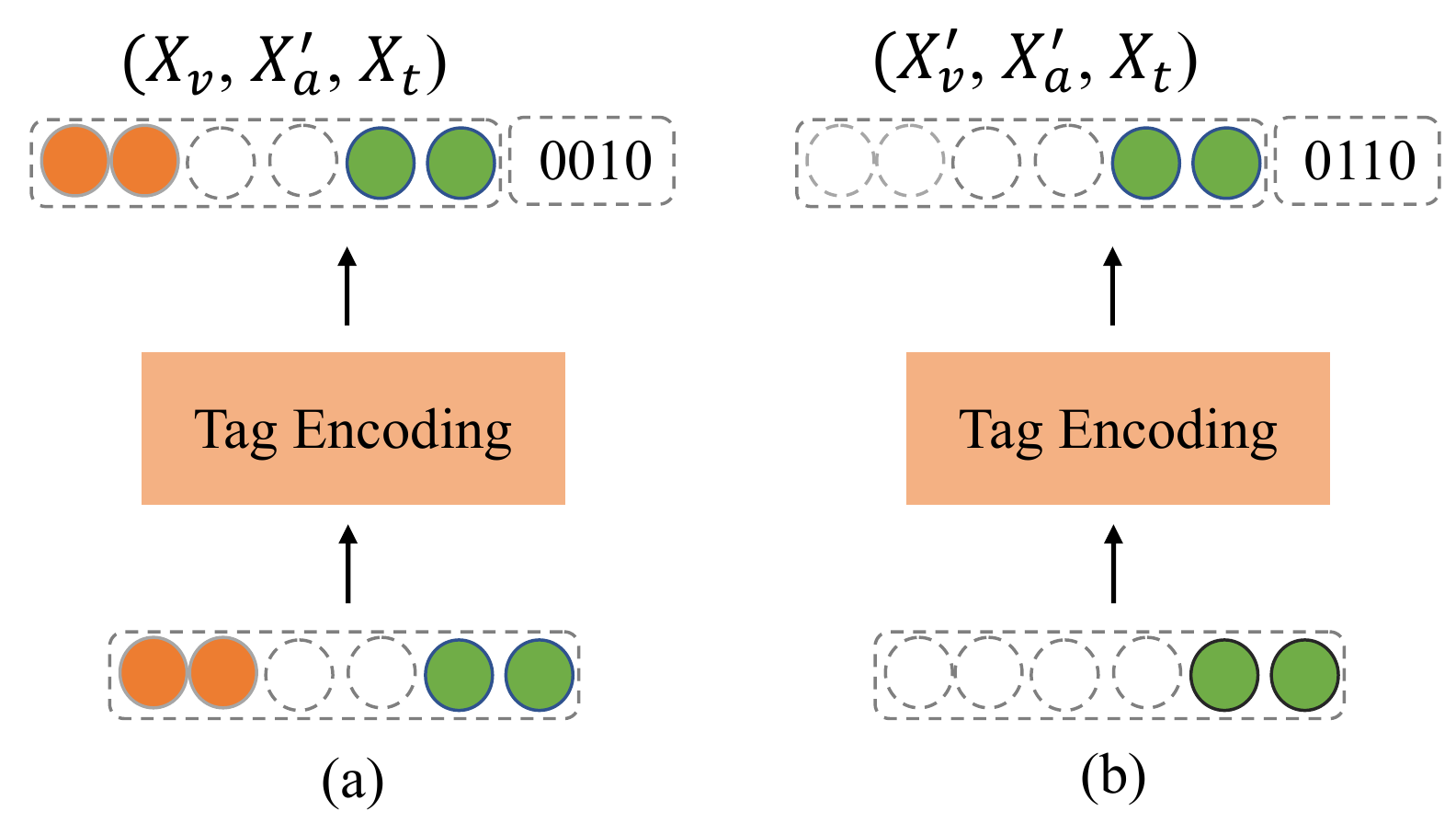}
    \caption{Examples of modality tags. (a) modality tag with one missing modality. (b) modality tag with two missing modalities.}
    \label{fig3}
\end{figure}

To specify uncertain missing modalities, we employ a tag encoding module to mark them, and direct network's attention to these disabled modalities. In our settings, we adopt 4 digits (``0'' or ``1'') to label missing modalities. If partial modalities of the input are missing, we set the first digit as ``0'', otherwise ``1''. Besides, the last three digits are used to mark the corresponding visual, acoustic and textual modalities. As can be seen in Fig.~\ref{fig3}, we give two examples about modality tags: in Fig.~\ref{fig3}a, the acoustic modality is missing, and the tag is set as ``0010"; for multiple modalities missing cases (Fig.~\ref{fig3}b), we set the tag as ``0110" to mark visual and acoustic modalities. The benefits are twofold: 1) the tag encoding module can cover both single and multiple modalities missing conditions; and 2) the encoded tags can complementarily assist in the learning of the joint representations. To simplify mathematical expression, we denote all tags as $E_{tag}$.

\subsection{Common Space Projection}

\begin{figure}[t]
    \centering
    \includegraphics[width=0.73\linewidth ]{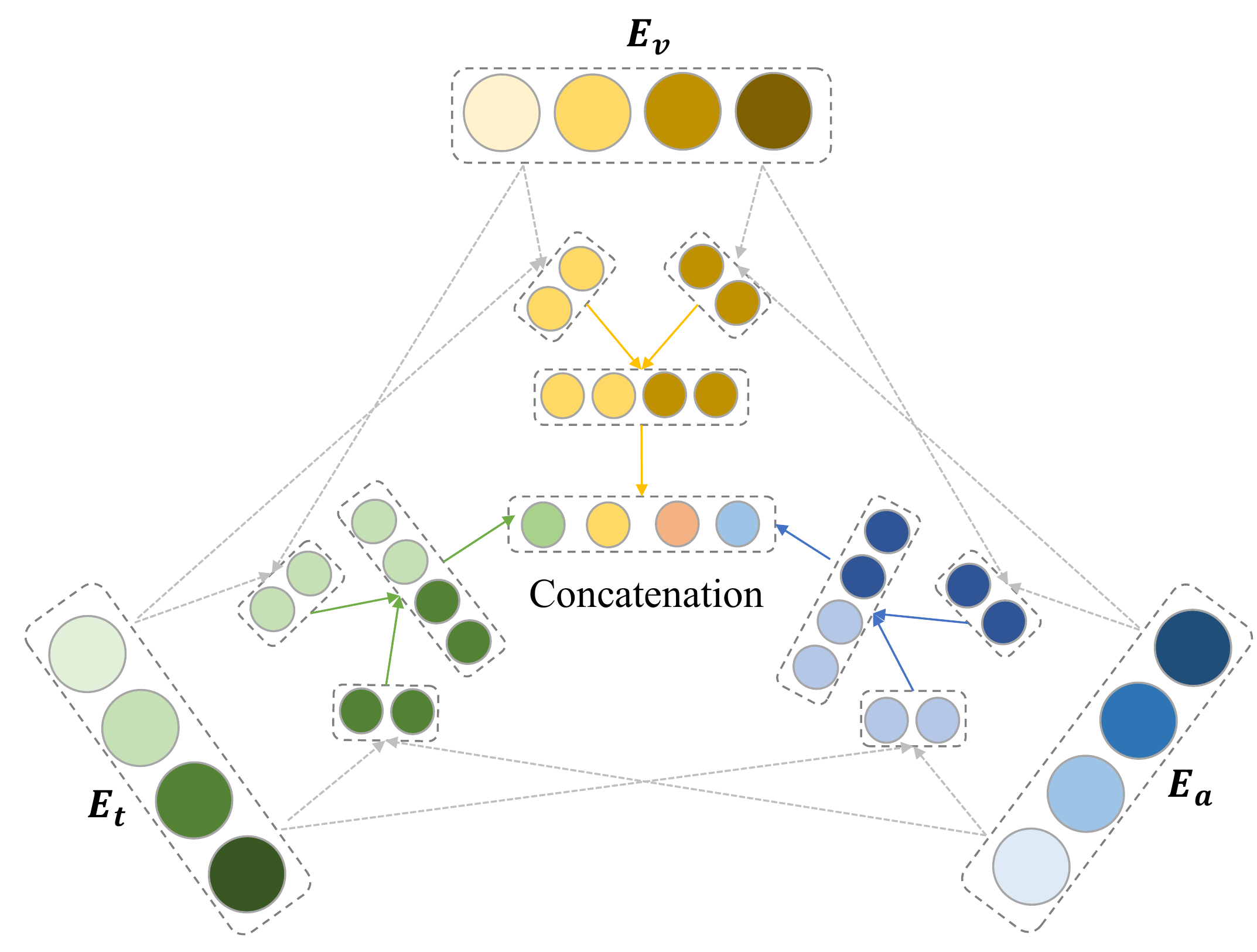}
    \caption{Illustration of the common space projection.}
    \label{fig4}
\end{figure}

After the tag encoding module, we now project three modalities into the common space. Previous works~\cite{xu2020reasoning, hazarika2020misa} that directly utilized simple feed-forward neural layers with same parameters for the projection, which may be failed when there are more than two modalities. To tackle the issue, we adopt a two-by-two projection pattern to acquire a more general space. As shown in Fig.~\ref{fig4}, for each single modality, we first obtain the self-related common space based on the following linear transformation:
\begin{equation}
    \begin{aligned}
        &C_{v} = [W_{va} E_{v} || W_{vt} E_{v}], \\
        &C_{a} = [W_{va} E_{a} || W_{ta} E_{a}], \\
        &C_{t} = [W_{vt} E_{t}\hspace{0.3em} || W_{ta} E_{t}], \\
    \end{aligned}
\end{equation}
where $W_{va}$, $W_{vt}$ and $W_{ta}$ are all weight matrices, and $||$ denotes the vertical concatenating operation. Then, we concatenate all common vectors and the encoded tag to eventually obtain the common joint representations $E_{all}$:
\begin{equation}
    E_{all} = [C_{v} || C_{a} || C_{t} || E_{tag}].
\end{equation}

\subsection{Transformer Encoder-Decoder}
To effectively model the long-term dependency of the intra-modal and the inter-modal information, we employ one sub-layer in Transformer~\cite{vaswani2017attention} to manage the information flow. As illustrated in Section~\ref{MultiHead}, the encoded outputs $E_{out}$ can be accessed by the multi-head attention and feed-forward networks:
\begin{equation}
    \begin{aligned}
        &E_{out} = MultiHead(E_{all}, E_{all}, E_{all}),\\
        &E_{out} = relu(E_{out}W_{e}^{1}+b_{e}^{1})W_{e}^{2}+b_{e}^{2},
    \end{aligned}
\end{equation}
where the query, the key, and the value are the same input $E_{all}$, $W_{e}^{1}$, and $W_{e}^{2}$ are two weight matrices, $b_{e}^{1}$ and $b_{e}^{2}$ are two learnable biases.

Similarly, the decoded outputs $D_{out}$ are formulated as follows:
\begin{equation}
    \begin{aligned}
        &D_{out} = MultiHead(E_{out}, E_{out}, E_{out}),\\
        &D_{out} = relu(D_{out}W_{o}^{1}+b_{o}^{1})W_{o}^{2}+b_{o}^{2},
    \end{aligned}
\end{equation}
where $W_{o}^{1}$, $W_{o}^{2}$, $b_{o}^{1}$, and $b_{o}^{2}$ are parameters.

\subsection{Training Objective}
The overall training objective ($\mathcal{L}_{total}$) is expressed as:
\begin{equation}
     \mathcal{L}_{total} =  \mathcal{L}_{cls} + \lambda_{1} \mathcal{L}_{forward} + \lambda_{2} \mathcal{L}_{backward} + \lambda_{3} \mathcal{L}_{tag},
    \label{loss}
\end{equation}
where $ \mathcal{L}_{cls}$ is the classification loss, $ \mathcal{L}_{forward}$ is the forward differential loss, $ \mathcal{L}_{backward}$ is the backward reconstruction loss, $ \mathcal{L}_{tag}$ is the tag recovery loss, and $\lambda_{1}$, $\lambda_{2}$ and $\lambda_{3}$ are the corresponding weights. We now introduce the loss terms in details.

\subsubsection{Forward Differential Loss ($ \mathcal{L}_{forward}$)}
As illustrated in Fig.~\ref{fig2}, the forward loss is calculated by the difference between the pre-trained output ($E_{pre}$) and the Transformer encoder output ($E_{out}$). Simiar to~\cite{zhao2021missing}, the pre-trained model is trained with full modality data, where features from three modalities are directly concatenated for classification. Thus, we employ the differential loss to guide the learning process for missing modalities. Specifically, the Kullback Leibler (KL) divergence loss is used:
\begin{equation}
    D_{KL}(p||q) = \sum_{i=1}^{N} p(x_{i}) \cdot \frac{p(x_{i})}{q(x_{i})},
\end{equation}
where $p$ and $q$ are two probability distributions. Since KL divergence is asymmetric, we adopt the Jensen-Shannon (JS) divergence loss instead:
\begin{equation}
    \begin{aligned}
         \mathcal{L}_{forward} &= JS(E_{out}||E_{pre}) \\
                       &= \frac{1}{2}(D_{KL}(E_{out}||E_{pre}) + D_{KL}(E_{pre}||E_{out})).
    \end{aligned}
\end{equation}

\subsubsection{Backward Reconstruction Loss ($ \mathcal{L}_{backward}$)}
For the backward loss, we aim to supervise the joint common vector reconstruction. Therefore, similar to the forward differential loss, we calculate the JS divergence loss between the Transformer decoder output ($D_{out}$) and the updated common joint representations ($E_{all}$):
\begin{equation}
    \begin{aligned}
         \mathcal{L}_{backward} &= JS(D_{out}||E_{all}) \\
                        &= \frac{1}{2}(D_{KL}(D_{out}||E_{all}) + D_{KL}(E_{all}||D_{out})).
    \end{aligned}
\end{equation}

\subsubsection{Tag Recovery Loss ($ \mathcal{L}_{tag}$)}
In our settings, the tag is attached to mark missing modalities, and we expect our network can pay more attention to them. To better guide the reconstruction of the attached tag, we design a tag recovery loss to direct the process. The reason why we choose the Mean Absolute Error(MAE) loss is that MAE is less sensitive to outliers with the absolute function. Thus, MAE is adopted to calculate the loss between $E_{tag}$ and the last four digits of $D_{out}$:
\begin{equation}
    \begin{aligned}
        & \mathcal{L}_{tag} = \frac{1}{N}\sum_{i=1}^{N}|E_{tag}^{i} - D_{tag}^{i}|,\\
        &  D_{tag} = Sigmoid(D_{out}[-4:]).
    \end{aligned}
\end{equation}

\subsubsection{Classification Loss ($ \mathcal{L}_{cls}$)}
For the final classification module, we feed $E_{out}$ into a fully connected network with the softmax activation function:
\begin{equation}
   P_{c} = softmax(W_{c}E_{out} + b_{c}),
\end{equation}
where $W_{c}$ and $b_{c}$ are the learned weights and bias. In detail, we employ the standard cross-entropy loss for this task, that is:
\begin{equation}
     \mathcal{L}_{cls} = -\frac{1}{N}\sum_{n=1}^{N}y_{n}log\hat{y}_{n},
\end{equation}
where $N$ is the number of samples, $y_{n}$ is the true label of the $n$-th sample, and $\hat{y}_{n}$ is the predicted label.

\section{Experiments}~\label{experiment}

\begin{table}[t]
    \centering
    \caption{Detailed parameter settings in all experiments.}
    \begin{tabular}{lcc}
      \hline
      Description & Symbol & Value\\
      \hline
      Batch size & $b$ & 32\\
      Epoch number & $e$ & 20\\
      Dropout rate & $p$ & 0.3 \\
      Hidden size & $d$ & 300\\
      Missing rate & $\eta$ & [0, 0.5] \\
      Learning rate & $lr$ & 0.001\\
      Maximum textual length & $n_{t}$ & 25 \\
      Maximum visual length & $n_{v}$ & 100 \\
      Maximum acoustic length & $n_{a}$ & 150 \\
      Loss weights & $\lambda_{1}, \lambda_{2}, \lambda_{3}$ & 0.1 \\
      \hline
    \end{tabular}
    \label{para}
\end{table}

All experiments are carried out on a Linux server (Ubuntu 18.04.1) with a Intel(R) Xeon(R) Gold 5120 CPU, 8 Nvidia 2080TI GPUs and 128G RAM. Datasets and experimental settings are described as follows:

\textbf{Datasets:} We conduct several experiments on CMU-MOSI~\cite{zadeh2016multimodal} and IEMOCAP~\cite{busso2008iemocap} datasets. Both datasets are multimodal benchmarks for sentiment recognition, including visual, textual, and acoustic modalities. For the CMU-MOSI dataset, it contains 2199 segments from 93 opinion videos on YouTube. The label of each sample is annotated with a sentiment score in [-3, 3]. Following Yu \textit{et al.}~\cite{yu2021learning}, we transform the score into negative, neutral and positive labels. For the IEMOCAP dataset, it contains 5 sessions, and each session contains about 30 videos, where each video contains at least 24 utterances. The annotated labels are: neutral, frustration, anger, sad, happy, excited, surprise, fear, disappointing, and other. Specifically, we report three-classes (negative: [-3,0), neutral:[0], positive: (0,3]) results on CMU-MOSI, and two-classes (negative:[frustration, angry, sad, fear, disappointing], positive:[happy, excited]) on IEMOCAP.

\textbf{Parameters:} Following standard methods, we tune our model using five-fold validation and grid-searching on the training set. The learning rate $lr$ is selected from $\{0.1, 0.001, 0.0005, 0.0001\}$, the batch size $b \in \{32, 64, 128\}$, and the hidden size $d \in \{64, 128, 300, \\768\}$. Adam~\cite{kingma2014adam} is adopted to minimize the total loss given in Eq.~(\ref{loss}). The epoch number is 20, the batch size is 32, the loss weight is set to 0.1, and the parameters are summarized in Table~\ref{para}.

\textbf{Evaluation Metric:} $Accuracy$ and $Macro-F1$ are used to measure the performance of the models, which are defined as follows:
\begin{equation}
    \begin{aligned}
        &Accuracy = \frac{T_{true}}{N},\\
        &F1 = \frac{2PR}{P+R},
    \end{aligned}
\end{equation}
where $T_{true}$ is the number of correctly predicted samples, $N$ is the total number of samples, $P$ is the positive predictive value, and $R$ is the recall value.

\subsection{Feature Extraction}
\noindent{\textbf{Visual Representations:}} The CMU-MOSI~\cite{zadeh2016multimodal} and IEMOCAP~\cite{busso2008iemocap} datasets mainly consist of human conversations, where visual features are mainly composed of human faces. Following~\cite{zadeh2018multimodal, yu2020ch}, we also adopt OpenFace2.0 toolkit~\cite{baltrusaitis2018openface} to extract facial features. Except for the first to the fifth columns data, we finally obtain 709-dimensional visual representations, where the face, the head, and the eye movement are included.

\noindent{\textbf{Textual Representations:}}
For each textual utterance, the pre-trained Bert~\cite{devlin2019bert} is utilized to extract textual features. Eventually, we adopt the pre-trained uncased BERT-base model (12-layer, 768-hidden, 12-heads) to acquire 768-dimensional word vectors.

\noindent{\textbf{Acoustic Representations:}}
As an audio analysis toolkit, Librosa~\cite{mcfee2015librosa} shows an excellent ability to extract acoustic features. For CMU-MOSI and IEMOCAP datasets, each audio is mixed to the mono and is re-sampled to 16000 Hz. Besides, each frame is separated by 512 samples, and we choose the zero crossing rate, the mel-frequency cepstral coefficients (MFCC) and the Constant-Q Togram (CQT) features to represent audio segments. Finally, we concatenate three features to yield 33-dimensional acoustic features.
\begin{table*}[t]
    \centering
     \caption{Results of all baselines of missing a single modality, where the best results are in bold.}
    \begin{tabular}{cccc cccc cccc cc}
    \hline
      \multirow{2}*{Datasets} & \multirow{2}*{Models} & \multicolumn{2}{c}{0} & \multicolumn{2}{c}{0.1} & \multicolumn{2}{c}{0.2} & \multicolumn{2}{c}{0.3} & \multicolumn{2}{c}{0.4} & \multicolumn{2}{c}{0.5}  \\
       &  & M-F1 & ACC  &  M-F1 & ACC  &  M-F1 & ACC  &  M-F1 & ACC  &  M-F1 & ACC  &  M-F1 & ACC \\
    \hline
        \multirow{6}*{CMU-MOSI}
          & AE  & 56.78 & 79.69 & 54.07 & 79.17 & 53.40 & 78.13 & 51.28 & 72.53 & 50.75 & 73.48 & 44.99 & 69.32 \\
          & CRA  & 56.85 & 79.73 & 54.37 & 79.38 & 53.57 & 78.24 & 51.67 & 72.84 & 51.02 & 73.79 & 45.38 & 69.45 \\
          & MCTN  & 57.32 & 79.75 & 55.48 & 79.87 & 53.99 & 77.49 & 52.31 & 71.59 & 51.64 & 73.81 & 45.76 & 68.11 \\
          & TransM  & 57.84 & 80.21 & 57.53 & 79.69 & 55.21 & 78.42 & 52.87 & 72.92 & 52.49 & 72.40 & 45.86 & 68.23 \\
          & MMIN  & \textbf{60.41} & 82.29 & 57.75 & 81.86 & 55.38 & 80.20 & 53.65 & 79.24 & 52.55 & 76.33 & 48.95 & 70.76 \\
          & Ours  & 58.32 & \textbf{84.90} & \textbf{58.21} & \textbf{84.46} & \textbf{55.46} & \textbf{81.25} & \textbf{55.11} & \textbf{80.73} & \textbf{54.11} & \textbf{80.21} & \textbf{51.71} & \textbf{74.04} \\
    \hline
    \multirow{6}*{IEMOCAP}
          & AE  & 76.15 & 82.09 & 75.24 & 80.26 & 75.02 & 78.01 & 73.92 & 77.43 & 70.19 & 76.01 & 67.27 & 76.43 \\
          & CRA  & 77.05 & 82.13 & 75.95 & 80.97 & 75.13 & 78.09 & 74.02 & 78.11 & 70.69 & 76.12 & 67.75 & 76.49 \\
          & MCTN  & 78.57 & 82.27 & 77.74 & 81.02 & 75.37 & 78.27 & 74.69 & 78.52 & 71.75 & 76.29 & 68.17 & 76.63 \\
          & TransM  & 79.57 & 82.64 & 78.03 & 81.86 & 76.33 & 80.43 & 75.83 & 78.64 & 72.01 & 77.27 & 68.57 & 76.65 \\
          & MMIN  & 80.83 & 83.43 & 78.85 & 82.58 & 77.09 & 81.27 & 76.63 & 80.43 & 72.81  & 78.43 & 70.58 & 77.45 \\
          & Ours  & \textbf{81.15} & \textbf{85.39} & \textbf{79.99} & \textbf{85.09} & \textbf{79.10} & \textbf{84.07} & \textbf{78.45} & \textbf{83.25} & \textbf{76.74} & \textbf{82.75} & \textbf{74.43} & \textbf{82.43}\\
    \hline
    \end{tabular}
    \label{tab_res1}
\end{table*}

\begin{table*}[t]
    \centering
    \caption{Results of all baselines of missing multiple modalities, where the best results are in bold.}
    \begin{tabular}{cccc cccc cccc cc}
    \hline
      \multirow{2}*{Datasets} & \multirow{2}*{Models} & \multicolumn{2}{c}{0} & \multicolumn{2}{c}{0.1} & \multicolumn{2}{c}{0.2} & \multicolumn{2}{c}{0.3} & \multicolumn{2}{c}{0.4} & \multicolumn{2}{c}{0.5}  \\
       &  & M-F1 & ACC  &  M-F1 & ACC  &  M-F1 & ACC  &  M-F1 & ACC  &  M-F1 & ACC  &  M-F1 & ACC \\
    \hline
        \multirow{6}*{CMU-MOSI}
          & AE      & 56.78 & 79.69 & 52.80 & 75.65 & 50.84 & 74.18 & 46.23 & 69.18 & 44.40 & 69.05 & 40.29 & 66.01 \\
          & CRA     & 56.81 & 79.72 & 52.85 & 75.68 & 51.02 & 74.73 & 46.87 & 69.23 & 45.17 & 69.48 & 41.77 & 66.82 \\
          & MCTN    & 56.85 & 79.73 & 52.97 & 75.89 & 51.75 & 74.16 & 46.98 & 69.29 & 45.73 & 69.55 & 42.98 & 67.02 \\
          & TransM  & 57.84 & 80.21 & 53.49 & 77.08 & 51.97 & 74.24 & 48.23 & 70.51 & 47.02 & 70.38 & 43.28 & 67.74 \\
          & MMIN  & \textbf{60.41} & 82.29 & 55.49 & 80.12 & 52.79 & 76.26 & 48.97 & 73.27 & 47.39 & 74.28 & 44.63 & 68.92 \\
          & Ours  & 58.32 & \textbf{84.90} & \textbf{56.38} & \textbf{81.77} & \textbf{54.87} & \textbf{81.07} & \textbf{52.12} & \textbf{77.60} & \textbf{51.19} & \textbf{76.56} & \textbf{51.15} & \textbf{73.23} \\
    \hline
    \multirow{6}*{IEMOCAP}
          & AE    & 76.15 & 82.09 & 75.07 & 79.84 & 74.20 & 76.91 & 71.55 & 76.07 & 69.73 & 75.16 & 67.15 & 75.22 \\
          & CRA  & 77.05 & 82.13 & 75.21 & 79.95 & 74.22 & 77.03 & 71.86 & 76.41 & 70.13 & 75.29 & 67.31 & 75.42 \\
          & MCTN  & 78.57 & 82.27 & 76.83 & 80.56 & 74.77 & 77.89 & 72.27 & 77.03 & 71.02 & 75.84 & 67.51 & 75.88 \\
          & TransM  & 79.57 & 82.64 & 77.21 & 81.13 & 75.87 & 79.01 & 72.36 & 78.15 & 71.38 & 76.88 & 68.02 & 76.04 \\
          & MMIN  & 80.83 & 83.43 & 78.02 & 82.32 & 76.38 & 79.53 & 73.05 & 79.02 & 71.22  & 77.27 & 69.39 & 77.01 \\
          & Ours  & \textbf{81.15} & \textbf{85.39} & \textbf{78.37} & \textbf{83.63} & \textbf{77.55} & \textbf{82.33} & \textbf{76.14} & \textbf{82.21} & \textbf{74.09} & \textbf{81.94} & \textbf{72.49} & \textbf{80.57}\\
    \hline
    \end{tabular}
    \label{tab_res2}
\end{table*}

\subsection{Baselines}
To evaluate the performance of our approach, the following baselines are chosen for comparison:

\textbf{AE~\cite{baldi2012autoencoders}:} An efficient data encoding network trained to copy its input to its output. In our implementation, we employ 5 AEs with each layer of the size [512, 256, 128, 64].

\textbf{CRA~\cite{tran2017missing}:} A missing modality reconstruction framework that employed the residual connection mechanism to approximate the difference between the input data. In our implementation, we add a residual connection for the input with the same layer setting in AE~\cite{baldi2012autoencoders}.

\textbf{MCTN$\footnote{https://github.com/hainow/MCTN}$~\cite{pham2019found}:} A method to learn robust joint representations by translating among modalities, claiming that translating from a source modality to a target modality can capture joint information among modalities.

\textbf{TransM~\cite{wang2020transmodality}:} An end-to-end translation based multimodal fusion method that utilized Transformer to translate among modalities and encoded multimodal features. In our implementation, we concatenate 6 MAE losses between two modalities transformation.

\textbf{MMIN$\footnote{https://github.com/AIM3-RUC/MMIN/tree/master}$~\cite{zhao2021missing}:} A unified multimodal emotion recognition model that adopted the cascade residual auto-encoder and cycle consistency learning to recover missing modalities.

\textbf{TATE:} Our proposed model.

\subsection{Overall Results}
For the single modality missing case, the experimental results are shown in Table~\ref{tab_res1}, where the missing ratio is set from 0 to 0.5. Specifically, we report triple classification results on CMU-MOSI and two classification results on IEMOCAP. With the increment of missing rate, the overall results present a descending trend. Except for the M-F1 value under the full modality condition is lower about 2.02\% than MMIN on the CMU-MOSI dataset, our proposed method achieves the best results on other settings, validating the effectiveness of our model. As can be seen in the table, compared to auto-encoder based methods (AE, CRA), translation-based methods (MCTN, TransM) achieve better performance, probably due to the fact that end-to-end translation among modalities can better fuse the multimodal information. Besides, the comparative experiments suggest that the backward decoder can assist the forward encoder, so as to further improve the overall performance.

For multiple modalities missing cases, we also present related findings in Table~\ref{tab_res2}. In this setting, one or two modalities are randomly discarded. It can be seen that our proposed model still improves about 0.89\% to 3.10\% on M-F1 and about 1.31\% to 4.81\% on ACC compared to other baselines, demonstrating the robustness of the network. Owing to the forward differential loss and the assistance of tag, our model can still capture semantic-relevant information. More comparison will be given in Section~\ref{sec_tag}.

\subsection{Ablation Study}
\begin{table*}[t]
    \centering
    \caption{Comparison of all modules in TATE.}
    \begin{tabular}{c cccc cccc cccc}
    \hline
          \multirow{2}*{Modules} & \multicolumn{2}{c}{0} & \multicolumn{2}{c}{0.1} & \multicolumn{2}{c}{0.2} & \multicolumn{2}{c}{0.3} & \multicolumn{2}{c}{0.4} & \multicolumn{2}{c}{0.5}  \\
          & M-F1 & ACC  &  M-F1 & ACC  &  M-F1 & ACC  &  M-F1 & ACC  &  M-F1 & ACC  &  M-F1 & ACC \\
    \hline
        V & 37.84 & 56.25 & - & - & - & - & - & - & - & - & - & - \\
        A & 39.82 & 59.90 & - & - & - & - & - & - & - & - & - & - \\
        T & 55.63 & 76.17 & - & - & - & - & - & - & - & - & - & - \\
      V+A & 40.71 & 61.26 & 38.93 & 59.10 & 38.07 & 56.94 & 37.42 & 56.18 & 36.98 & 55.43 & 36.66 & 54.17 \\
      V+T & 56.98 & 79.13 & 56.41 & 78.67 & 55.07 & 76.29 & 54.83 & 74.90 & 52.86 & 74.15 & 50.32 & 72.92 \\
      A+T & 57.69 & 80.65 & 57.01 & 79.47 & 55.23 & 77.44 & 55.02 & 75.39 & 53.92 & 74.37 & 51.25 & 73.13 \\
    V+A+T & \textbf{58.32} & \textbf{84.90} & \textbf{58.21} & \textbf{84.46} & \textbf{55.46} & \textbf{81.25} & \textbf{55.11} & \textbf{80.73} & \textbf{54.11} & \textbf{80.21} & \textbf{51.71} & \textbf{74.04} \\
    \hline
    -w/o tag & 57.95 & 80.21 & 57.86 & 80.99 & 54.83 & 79.90 & 53.71 & 79.89 & 52.59 & 76.75 & 49.17 & 72.05 \\
    -w/o tag loss & 58.04 & 82.81 & 57.92 & 81.98 & 55.32 & 80.77 & 53.95 & 80.28 & 52.83 & 77.01 & 49.41 & 72.92 \\
    -w/o forward loss & 52.39 & 76.21 & 51.83 & 75.52 & 50.16 & 73.39 & 49.87 & 72.15 & 48.26 & 71.12 & 47.29 & 70.35 \\
    -w/o backward loss & 53.85 & 77.43 & 52.88 & 77.08 & 51.85 & 74.09 & 51.07 & 74.21 & 49.01 & 71.92 & 48.53 & 71.08\\
    -w/o common space & 54.03 & 79.76 & 53.20 & 77.59 & 52.97 &75.99 & 51.23 & 75.01 & 49.83 & 72.37 & 49.05 & 71.85 \\
    \hline
    \end{tabular}
    \label{tab_abs}
\end{table*}
To explore the effects of different modules in TATE, we evaluate our model with several settings: 1) using only one modality; 2) using two modalities; 3) removing the tag encoding module; 4) removing the common space projection module; 5) removing the tag recovery loss; 6) removing the forward differential loss; and 7) removing the backward reconstruction loss.

According to Table~\ref{tab_abs}, one interesting finding is that the performance drops sharply when the textual modality is missing, validating that textual information dominates in the multimodal sentiment analysis. A possible explanation for these results is that textual information is the manual transcription. However, similar reductions are not observed when removing the visual modality. We conjecture that the visual information is not well extracted due to the minor changes to the face. Besides, the top half of the table shows that the combination of two modalities provides better performance than single modality, indicating that two modalities can learn complementary features. As for the effects of different modules, the performance of the forward differential module decreases about 4.52\% to 6.38\% on M-F1 and about 3.69\% to 9.09\% on ACC compared to the whole model, demonstrating the importance of the forward guidance. Since we employ full modality to pre-train the guidance network, the forward JS divergence loss serves as a good supervision. One striking result to emerge from this table is that the tag encoding module slightly improves the performance as expected. To further validate the effectiveness of the tag encoding module, we conduct several experiments in the following sub-section.

\subsection{Effects of the Tag Encoding}\label{sec_tag}
\begin{table}[h]
    \centering
    \caption{Improvements of the Tag Encoding.}
    \begin{tabular}{c cc cc}
    \hline
        \multirow{2}*{Model} & \multicolumn{2}{c}{Basic} & \multicolumn{2}{c}{+Tag}  \\
                             & M-F1 & ACC & M-F1 & ACC \\
    \hline
                       AE      & 51.28 & 72.53 & 53.25 (3.69\% $\uparrow$) & 75.21 (3.56\% $\uparrow$) \\
                       TransM  & 52.87 & 72.92 & 54.79 (3.50\% $\uparrow$) &  76.02 (4.08\% $\uparrow$) \\
                       Ours    & 53.71 & 79.89 & 55.11 (2.54\% $\uparrow$) & 80.73 (1.04\% $\uparrow$)\\
    \hline
    \end{tabular}
    \label{tab_tag}
\end{table}

We incorporate the tag encoding module with two basic models: AE and TransM. The reason why we choose the above two models is that AE and TransM are two different kinds of encoders: AE is the auto-encoder based method, and TransM is the Transformer based method. For the above two models, we add tags after the feature extraction module. Table~\ref{tab_tag} presents the detailed results on the CMU-MOSI dataset with a 30\% missing rate. It can be seen that models with the tag encoding module improves about 2.54\% to 3.69\% on M-F1 and about 1.04\% to 4.08\% on ACC compared to basic models, showing the effectiveness of the tag encoding module. Owing to the added tag, the network can be better guided, and can further focus on missing modalities.

\begin{figure}[h]
    \centering
    \begin{minipage}[t]{0.48\linewidth}
        {\includegraphics[width=\linewidth ]{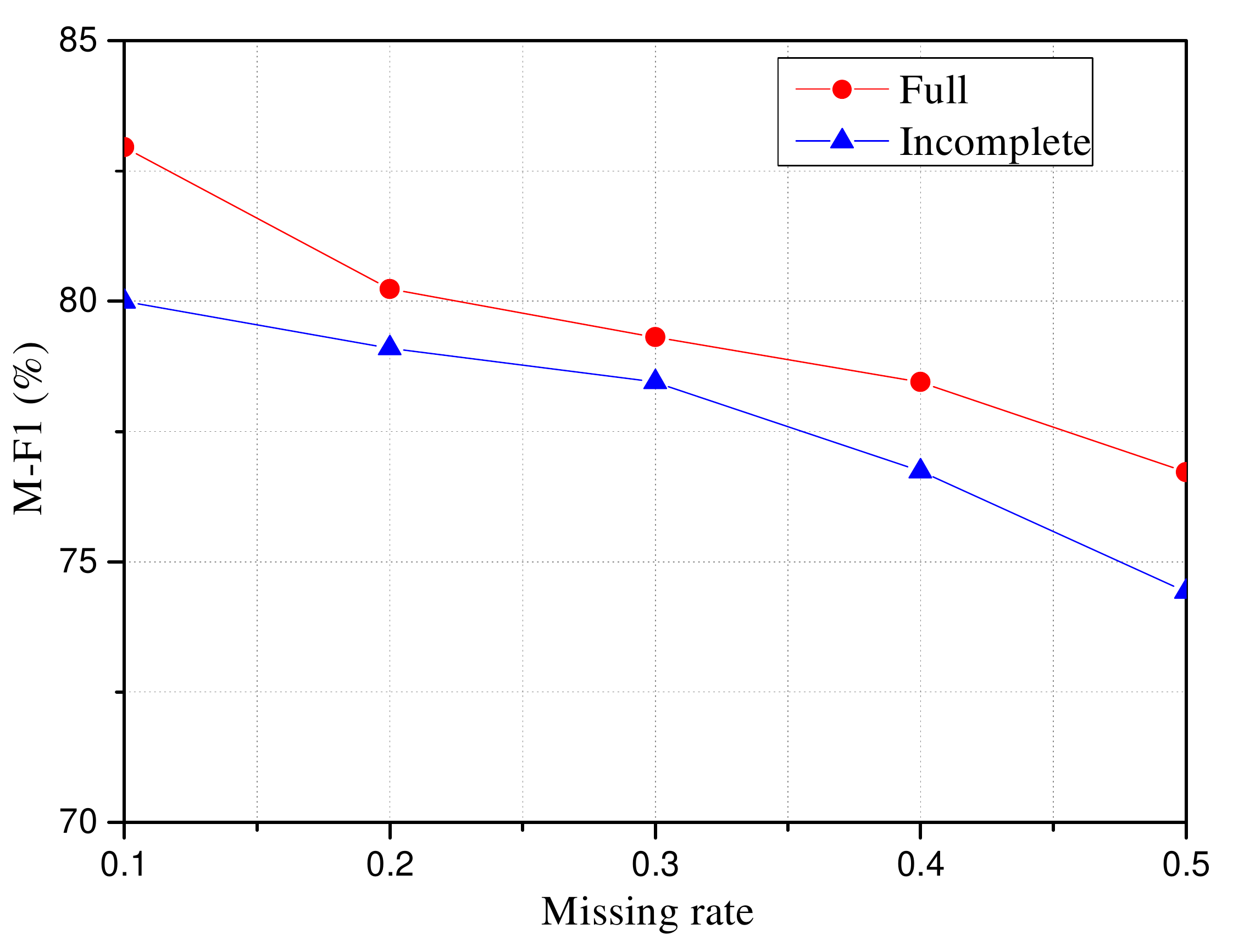}}
    \end{minipage}
    \begin{minipage}[t]{0.48\linewidth}
        {\includegraphics[width=\linewidth ]{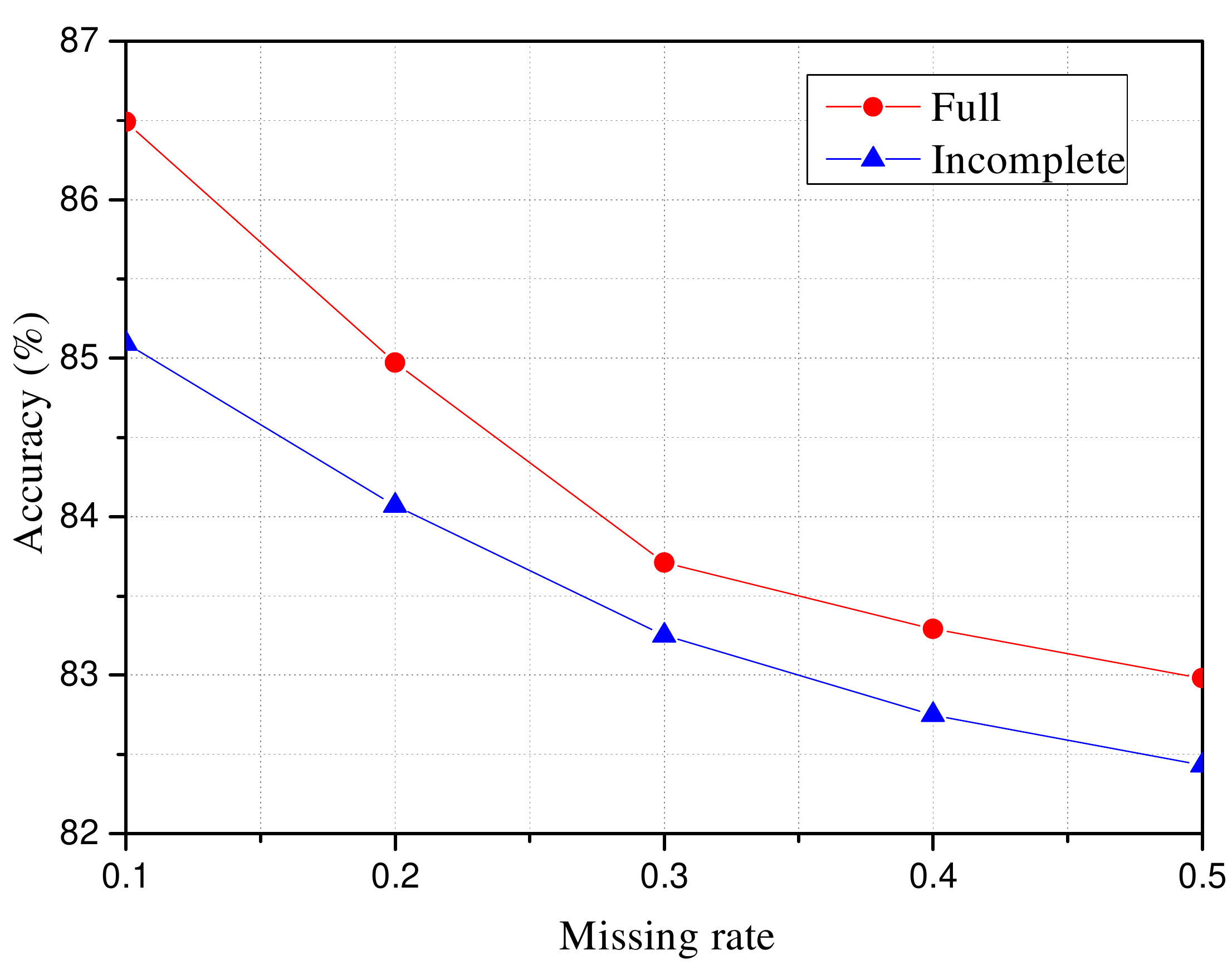}}
    \end{minipage}
    \caption{Comparison of full modalities and incomplete modalities during testing. (a) M-F1 values. (b) Accuracy values. }
    \label{fig5}
\end{figure}
\begin{figure*}[h]
    \centering

    \subfloat[$\eta = 0$]{\includegraphics[width=0.3\linewidth ]{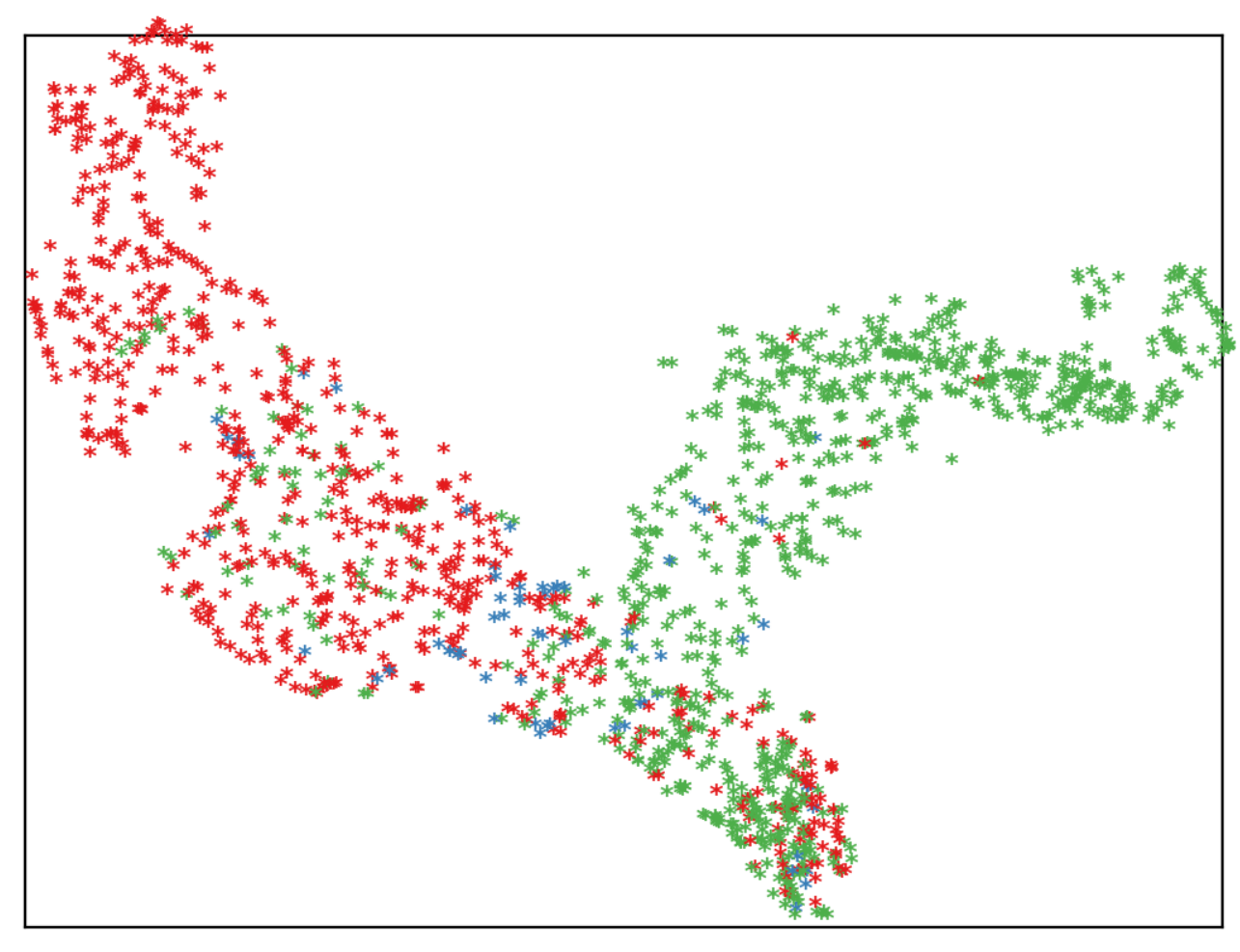}}\qquad
    \subfloat[$\eta = 0.1$]{\includegraphics[width=0.3\linewidth ]{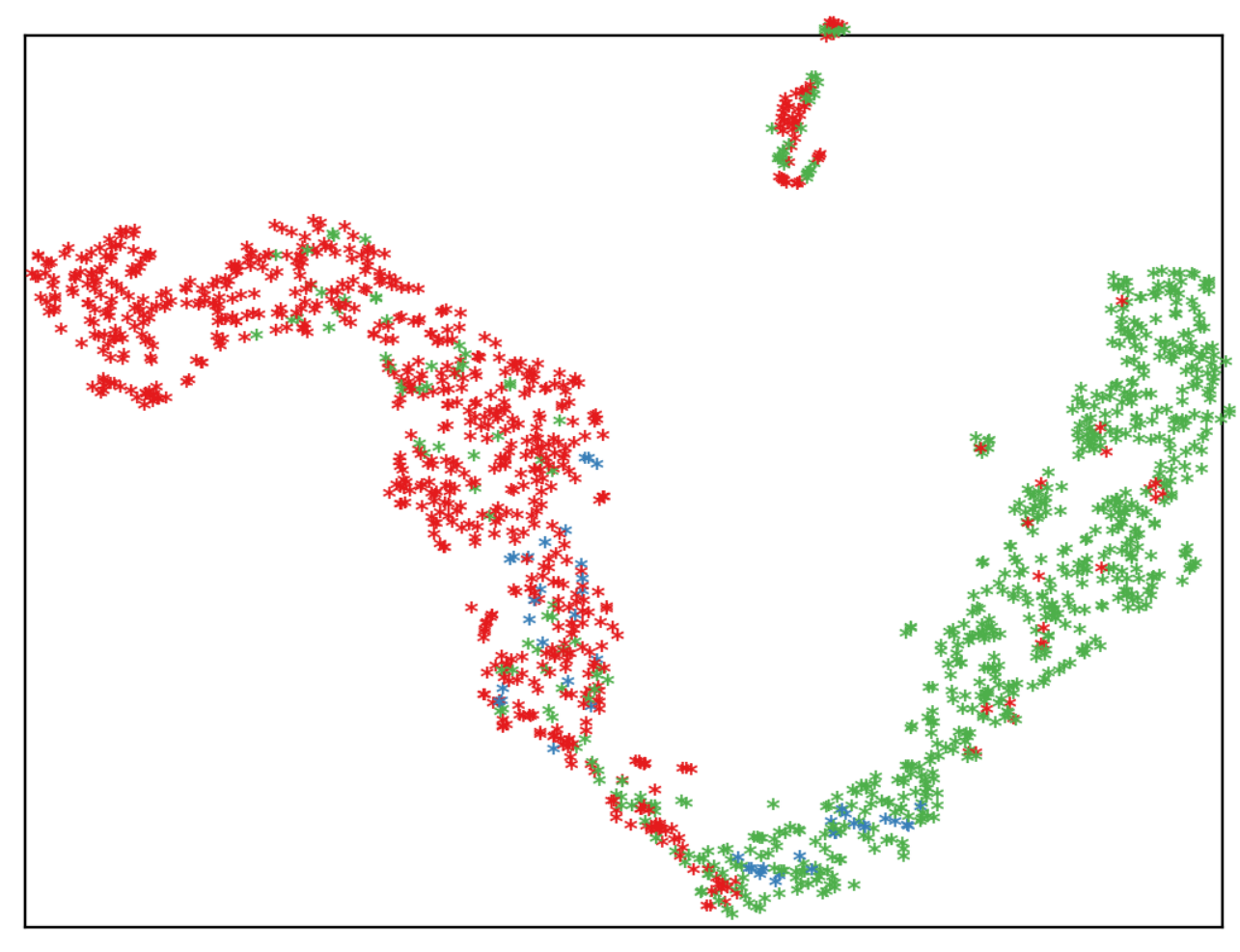}}\qquad
    \subfloat[$\eta = 0.2$]{\includegraphics[width=0.3\linewidth ]{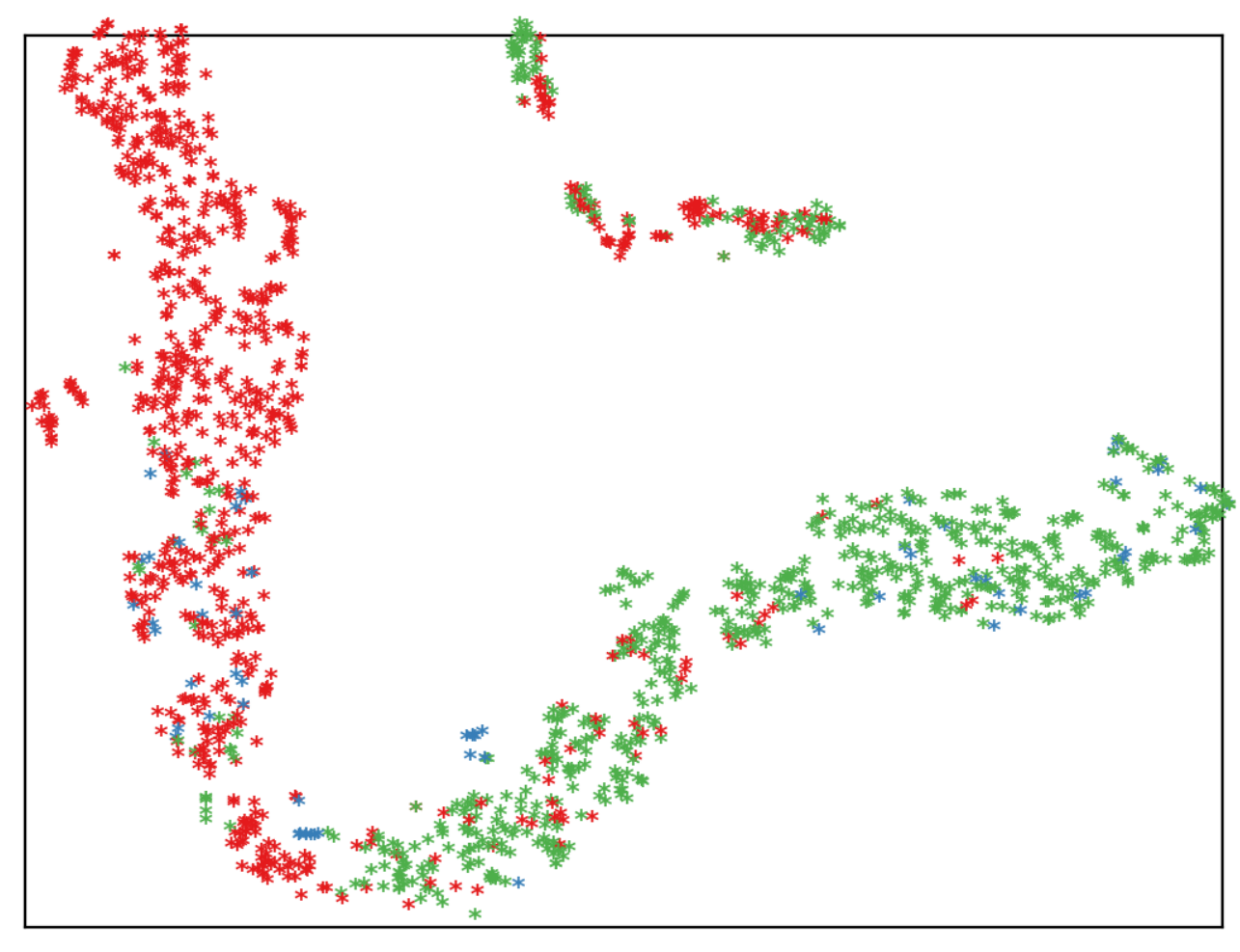}}
    \\
    \subfloat[$\eta = 0.3$]{\includegraphics[width=0.3\linewidth ]{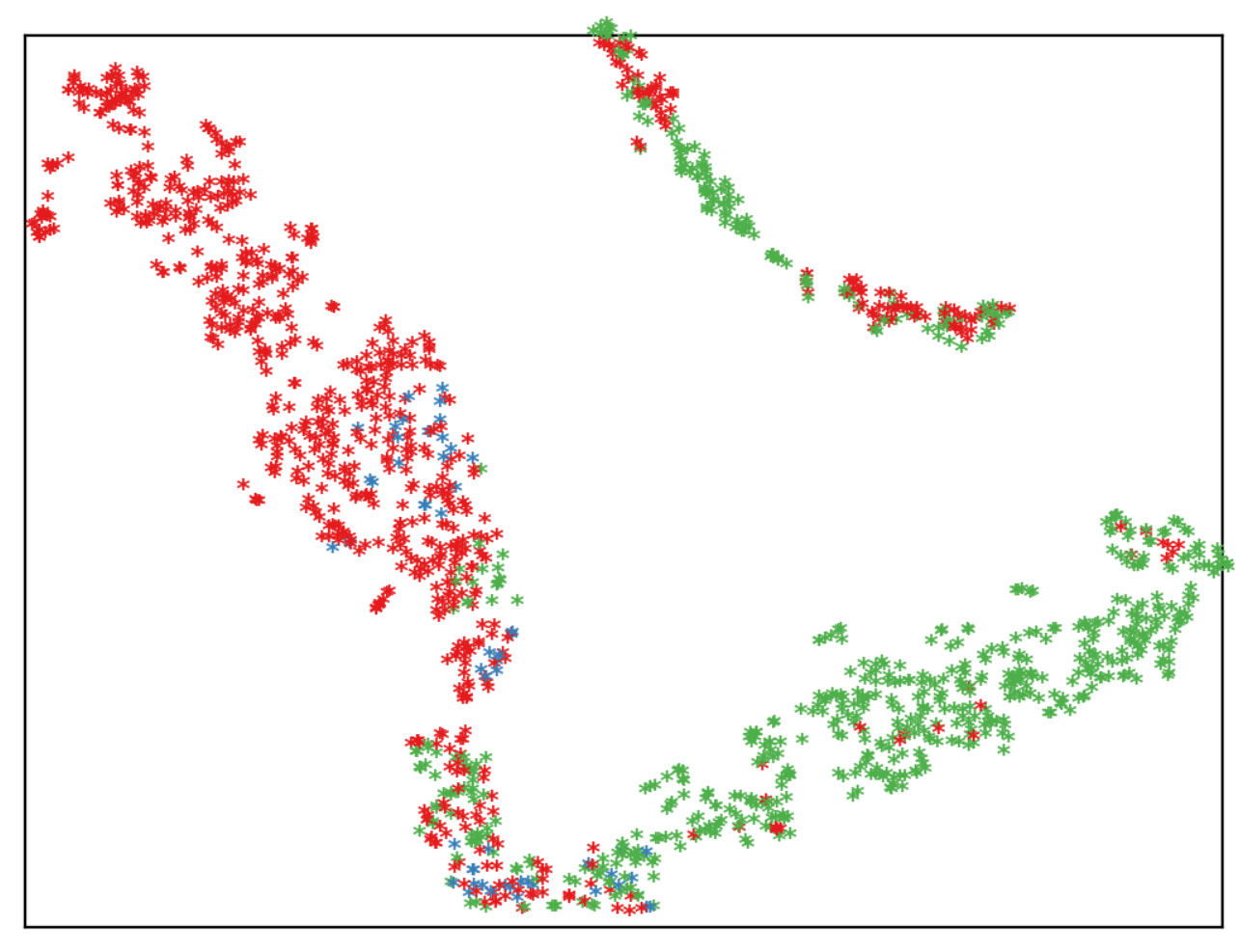}}\qquad
    \subfloat[$\eta = 0.4$]{\includegraphics[width=0.3\linewidth ]{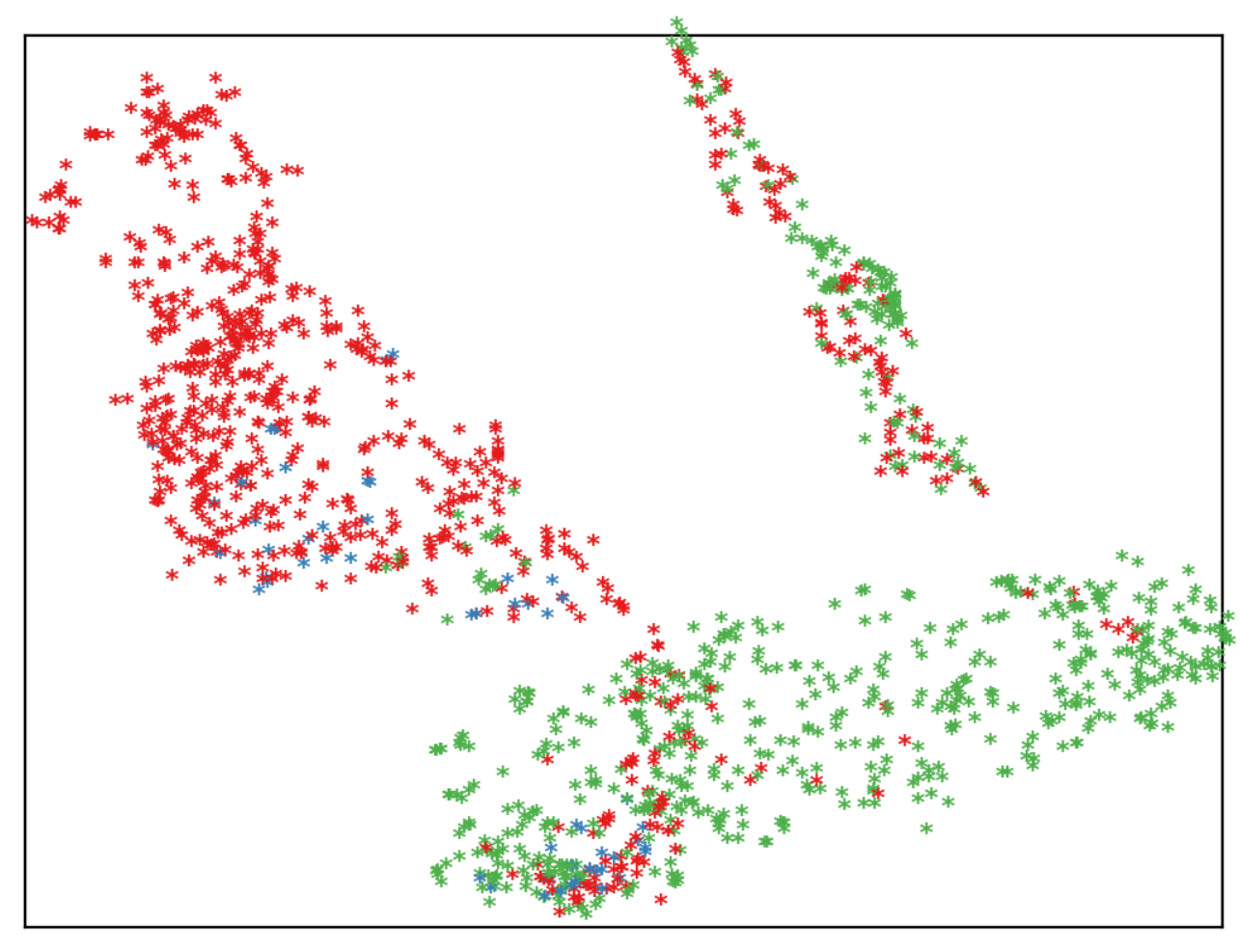}}\qquad
    \subfloat[$\eta = 0.5$]{\includegraphics[width=0.3\linewidth ]{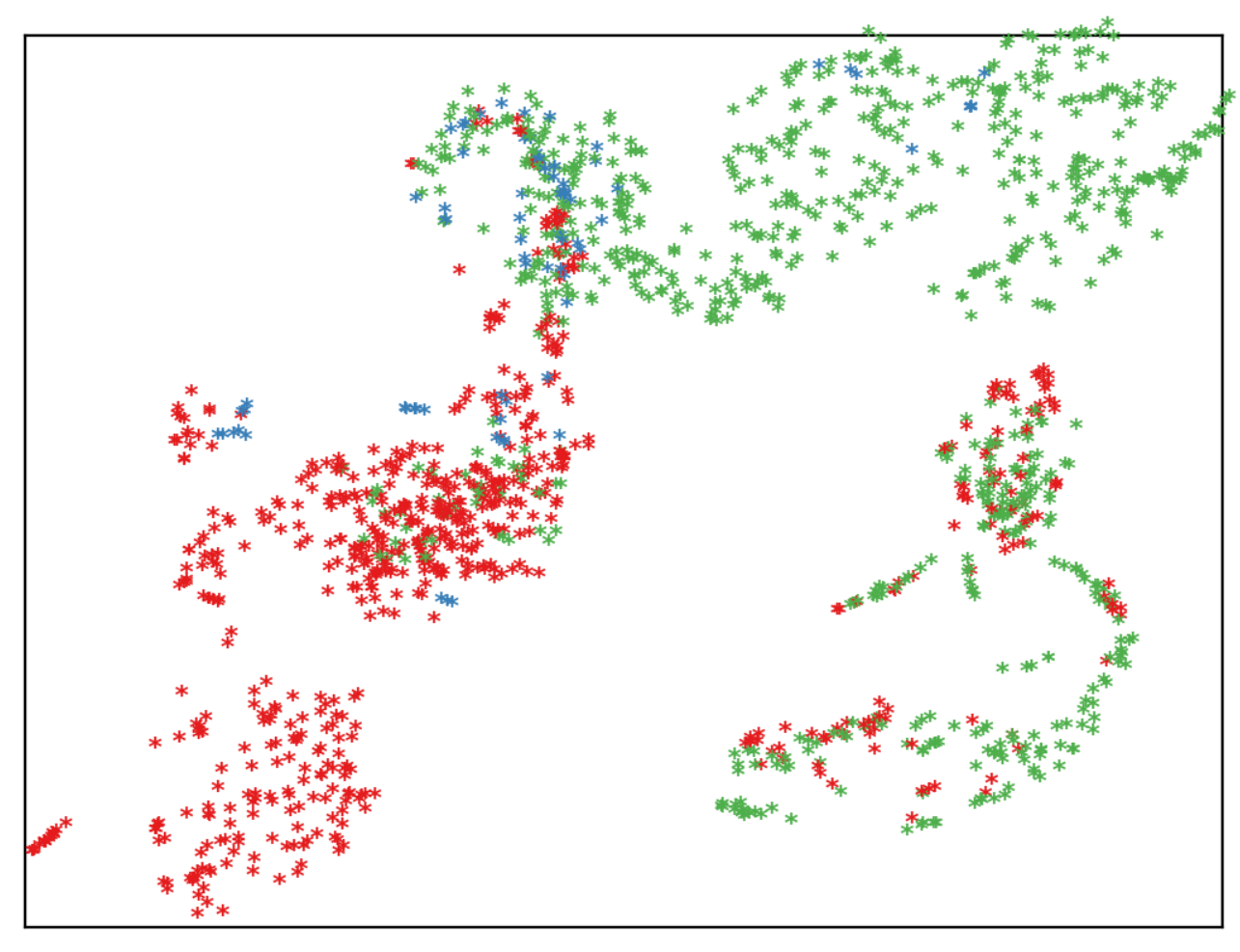}}

    \caption{Virtualization of joint representations with different rates of missing modalities (red: negative, blue: neutral, green: positive). (a) Full modalities, (b) missing rate 0.1, (c) missing rate 0.2, (d) missing rate 0.3, (e) missing rate 0.4, and (f) missing rate 0.5. }
    \label{fig6}
\end{figure*}
\subsection{Effects of the complete modality}
To see the difference between the complete and incomplete modalities of the test data, we first train the model with incomplete data, and then test the model with both full modality data and different missing rates of incomplete data. All experiments share the same parameters on the IEMOCAP dataset for a fair comparison. As can be seen in Fig.~\ref{fig5}, the gaps between two settings on M-F1 and ACC reach the minimum when the missing rate is 0.3. As the number of missing samples in the training data increases, the correlation among modalities becomes harder to capture, resulting in weaker test performance. However, the gap increases when the missing rate is bigger than 0.3. One possible explanation for the above results is that the model cannot learn the joint representation well because there are too many absent samples.

\subsection{Multi-classes on IEMOCAP}
\begin{table}[h]
    \centering
    \caption{Detailed distributions on IEMOCAP.}
    \resizebox{\linewidth}{10mm}{
    \begin{tabular}{clcc cccc c}
    \hline
        \multicolumn{2}{c}{Category} & Hap. & Ang. &  Sad. & Neu. & Fru. & Exc. & Sur\\
    \hline
        \multirow{2}*{4-calsses} & Train & 477 & 879 & 868 & 1385 & - & - & - \\
                                 & Test & 118 & 224 & 216 & 323 & - & - & - \\
    \hline
        \multirow{2}*{7-calsses} & Train & 476 & 891 & 873 & 1348 & 1458 & 848 & 87 \\
                                 & Test & 119 & 212 & 211 & 360 & 391 & 193 & 20 \\
    \hline
    \end{tabular}}
    \label{tab_iemo}
\end{table}

\begin{table}[h]
    \centering
    \caption{Results of multi-classes on IEMOCAP.}
    \setlength{\tabcolsep}{2.7mm}{
    \begin{tabular}{cccc cccc}
    \hline
        \multirow{2}*{Ratio} & \multicolumn{2}{c}{2-classes}  & \multicolumn{2}{c}{4-classes}  & \multicolumn{2}{c}{7-classes} \\
                             & M-F1 & ACC & M-F1 & ACC & M-F1 & ACC\\
    \hline
                          0  & 81.15 & 85.39 & 48.29 & 59.04 & 36.83 & 47.41 \\
                        0.1  & 79.99 & 85.09 & 46.56 & 57.70 & 36.01 & 45.73 \\
                        0.2  & 79.10 & 84.07 & 46.07 & 56.93 & 35.48 & 44.17 \\
                        0.3  & 78.45 & 83.25 & 45.69 & 56.27 & 35.21 & 43.02 \\
                        0.4  & 76.74 & 82.75 & 45.02 & 55.31 & 33.79 & 42.31 \\
                        0.5  & 74.43 & 82.43 & 44.15 & 54.85 & 33.28 & 42.08 \\
    \hline

    \hline
    \end{tabular}
    }
    \label{tab_iemo_res}
\end{table}

We also explore the performance of multiple classes on the IEMOCAP dataset. Apart from the two-classes results, we also choose happy, angry, sad and neutral emotions as the 4-classes experiment, and then choose the extra frustration, excited, and surprise emotions as the 7-classes experiment. The detailed distributions and results are presented in Table~\ref{tab_iemo} and Table~\ref{tab_iemo_res} respectively. It can be seen that both M-F1 value and ACC decrease with the increment of class numbers. By comparing the results with different rates of missing modalities, the gaps among 7-classes are smaller than that among 2-classes and 4-classes. Besides, closer inspection of Table~\ref{tab_iemo_res} shows that the overall performance drops sharply when the class number is 7, which is caused by the confusion of multiple classes, resulting in the difficulties in convergence of the model.

\subsection{Effects of different losses}
\begin{table}[h]
    \caption{Results of different losses.}
    \centering
    \resizebox{\linewidth}{19mm}{
    \begin{tabular}{cccc cccc}
         \hline
         \multirow{2}*{Datasets} & \multirow{2}*{Loss} & \multicolumn{2}{c}{0} & \multicolumn{2}{c}{0.2} & \multicolumn{2}{c}{0.4}  \\
         &  & M-F1 & ACC & M-F1 & ACC & M-F1 & ACC \\
         \hline
         \multirow{4}*{CMU-MOSI} & Cosine & 55.87 & 81.21 & 52.28 & 76.95 & 50.23 & 74.18 \\
                                 & MAE    & 56.21 & 82.05 & 52.37 & 77.16 & 51.15 & 74.76 \\
                                 & JS     & 57.89 & 84.15 & 53.42 & 79.05 & 52.17 & 75.22 \\
                                 & ours   & 58.32 & 84.90 & 55.46 & 81.25 & 54.11 & 80.21 \\
        \hline
         \multirow{4}*{IEMOCAP} & Cosine & 79.86 & 84.07 & 77.21 & 81.69 & 75.48 & 80.06 \\
                                & MAE    & 80.12 & 84.21 & 77.67 & 82.53 & 75.89 & 80.21 \\
                                & JS     & 80.73 & 85.03 & 78.56 & 83.21 & 76.34 & 81.07 \\
                                & ours   & 81.15 & 85.39 & 79.10 & 84.07 & 76.74 & 82.75 \\
         \hline
    \end{tabular}}
    \label{tab_loss}
\end{table}
To investigate the effects of different losses, we replace different loss function to see the performance. In detail, the cosine similarity loss, the MAE loss, and the JS divergence loss are chosen for comparison. We evaluate our model with 4 settings: 1) using the cosine similarity loss for $\mathcal{L}_{forward}$, $\mathcal{L}_{backward}$ and $\mathcal{L}_{tag}$; 2) using the MAE loss for $\mathcal{L}_{forward}$, $\mathcal{L}_{backward}$ and $\mathcal{L}_{tag}$; 3) using the JS divergence loss for$\mathcal{L}_{forward}$, $\mathcal{L}_{backward}$ and $\mathcal{L}_{tag}$; and 4) using the JS divergence loss for $\mathcal{L}_{forward}$ and  $\mathcal{L}_{backward}$, and using the MAE loss for $\mathcal{L}_{tag}$ (ours).

As can be seen in the Table~\ref{tab_loss}, our method achieves the best performance compared to other three loss settings on two datasets, showing the superiority of our model. Relatively, the results of applying JS divergence achieve secondary performance. Since the tag is composed of 4 digits (``0'' or ``1''), the MAE loss is more straightforward than JS divergence loss. Further analysis of the table suggests that the combination of the JS divergence loss and the MAE loss is beneficial in improving the overall performance.

\subsection{Visualization}
To better understand the learning ability of our model, we adopt the T-SNE toolkit~\cite{maaten2008visualizing} to visualize the joint representations under different rates of missing modalities. Specifically, we visualize about 1500 vectors learned by the Transformer encoder on the CMU-MOSI dataset, where the red, the blue, and the green color denote negative, neutral and positive respectively.

As shown in Figs.~\ref{fig6}(a)-(e), the overall joint representations obtain the similar distribution as the full modality condition. The majority of vectors are generally divided into three categories, where neutral samples is harder to classify because of their uncertain semantic. Besides, with the increment of missing ratio, the distributions become more discrete, especially when the missing ratio is bigger than 0.3. Apart from that, as can be seen in the top right-hand corner of Figs.~\ref{fig6}(b)-(e), the larger ratio of missing modalities, the wider outliers. The reason is that the model cannot converge with too many absent samples. While in Fig.~\ref{fig6}(f), the decision boundary is closer to the outliers when there are nearly half of missing samples. We suspect that absent samples dominate when training the model, resulting in a quite distinct distribution.

\section{Conclusion}~\label{conclusion}
In this paper, we propose a Tag-Assisted Transformer Encoder (TATE) network to handle the problem of missing partial modalities. Owing to the tag encoding technique, the proposed model can cover all uncertain missing cases, and the designed tag recovery loss can in turn supervise joint representation learning. Besides, more general aligned vectors are obtained by the common space module, and then are fed into the Transformer encoder for further process. At last, the final objective loss further directs the learning of missing modalities. All experimental results are conducted on CMU-MOSI and IEMOCAP datasets, showing the effectiveness of the proposed method.

In the future, this study may be further improved in the following ways: 1) for the common space projection module, we will try more fusion methods (e.g. add weights) to concatenate common feature vectors; and 2) for the Transformer encoder-decoder module, we employ the original sub-layer in Transformer as the basic semantic encoder. We attempt to adopt different structures of Transformer (e.g. Transformer-XL~\cite{dai2019transformer}, Reformer~\cite{kitaev2020reformer}, etc.) to observe the performance.


\begin{acks}
This work was supported by Macau Science and Technology Development Fund under SKL-IOTSC-2021-2023, 0072/2020/AMJ, 0015/-2019/AKP, 060/2019/A1, and 077/2018/A2, by Research Committee at University of Macau under MYRG2018-00029-FST and MYRG2019-00023-FST, and by Natural Science Foundation of China under 61971476.
\end{acks}

\bibliographystyle{ACM-Reference-Format}
\bibliography{ref}

\end{document}